\documentclass{article}
\usepackage{acra}
\usepackage{graphicx}
\usepackage{float}
\usepackage{amsmath}
\usepackage{booktabs}
\usepackage{float}
\usepackage{subcaption}
\usepackage{xcolor}
\usepackage{balance}

\usepackage{fancyhdr}
\fancypagestyle{acraheader}{
    \fancyhf{}

    \fancyhead[L]{
        \makebox[0pt][l]{
            \raisebox{-27pt}[0pt][0pt]{
                \parbox[b]{\textwidth}{
                    \footnotesize
                    \textbf{Cite as: Zhuoran Liang and Francisco Cruz.
                    Broad-Persistent Advising in Interactive
                    Reinforcement Learning via Latent State Encoding.
                    \textit{Proceedings of the Australasian Conference
                    on Robotics and Automation (ACRA), 2026.}}
                    \par\vspace{10pt}
                    \hrule height 0.4pt
                }
            }
        }
    }

}

\title{Broad-Persistent Advising in Interactive Reinforcement Learning via Latent State Encoding}

\author{
Zhuoran Liang$^{1}$ \qquad Francisco Cruz$^{1,2}$ \\
$^{1}$School of Computer Science and Engineering,
UNSW Sydney, Australia \\
$^{2}$Escuela de Ingeniería,
Universidad Central de Chile, Santiago, Chile \\
\texttt{jackliangzhuoran@gmail.com, f.cruz@unsw.edu.au}
}

\begin{document}

\maketitle
\thispagestyle{acraheader}

\begin{abstract}
Deep Interactive Reinforcement Learning (Deep IntRL) accelerates reinforcement learning training by incorporating external advice. However, conventional Deep IntRL uses each piece of advice only in the current step and discards it after a single interaction. Persistent Advising and Broad-Persistent Advising (BPA) extend this process by storing advice in persistent memory and reusing it in similar states through state generalisation based on K-means clustering. This research proposes Encoder BPA, which employs a latent encoder to retrieve and reuse advice. In this framework, a trained encoder is used to map a state sequence into a latent representation at every time step. Advice can be retrieved through latent comparison, together with a top-(k) aggregation and a consistency condition to improve reuse stability. Experimental results show that Encoder BPA improves learning efficiency and advice utilisation at an early stage of training compared with different Deep IntRL approaches, especially in complex and partially observable environments.
\end{abstract}

\section{Introduction}

The development of artificial intelligence has affected many industries, and technologies such as robotics and autonomous systems have become increasingly important. As a branch of AI techniques, reinforcement learning has attracted growing attention because it lets an agent learn, become familiar with, and work in an environment through interaction.

Reinforcement Learning (RL) describes a learning process in which an agent improves its decision-making through repeated interaction with an environment~\cite{suttonReinforcementLearningSecond2018,plaatDeepReinforcementLearning2022}. The agent can be understood as an autonomous system, such as a robot, a robotic arm, or a self-driving vehicle, while the environment refers to the surrounding space. As reinforcement learning has been applied to increasingly complex tasks, interactive reinforcement learning has emerged as an approach that incorporates external guidance into the learning process~\cite{cruzSurveyInteractiveReinforcement2021,millanHumanFeedbackContinuous2019}. This framework can be used in human-in-the-loop reinforcement learning tasks where humans can intervene in agent decision-making. Building on interactive reinforcement learning, the Persistent Advising and Broad-Persistent Advising (BPA) approach uses persistent memory to store and reuse advice from the external trainer. Rather than treating each piece of advice as a one-time interaction, BPA allows the agent to retain useful guidance and apply it again when it encounters the same or similar situations~\cite{nguyenBroadPersistentAdvisingApproach2023,bignoldHumanEngagementProviding2023}.

This research is motivated by the limitations of the BPA approach. In the original BPA approach, states are mapped into a small number of clusters using K-means clustering and previous advice can be used when the agent encounters another state assigned to the same cluster~\cite{nguyenBroadPersistentAdvisingApproach2023}. Although this provides efficient state generalisation, it may be coarse for high-dimensional robotic environments. For example, in the Domestic Robot Environment experiment of the original BPA study, each state is represented by a $64 \times 64$ RGB image, corresponding to $64 \times 64 \times 3$ input dimensions. At the same time, the elbow method selects only four clusters in total for the K-means clustering~\cite{nguyenBroadPersistentAdvisingApproach2023}. A limited number of clusters may force dissimilar states into the same cluster. As a result, the agent may reuse the same advice for behaviourally different states. In addition, the K-means clustering may be affected by the curse of dimensionality in a high-dimensional state space because of its distance measurement~\cite{murphyProbabilisticMachineLearning2022}.

To address these issues, this research proposes replacing the clustering-based state generalisation in the original BPA approach with a representation learning mechanism in which a neural network encoder learns latent representations from state histories and reuses previous advice via latent space advice retrieval. Based on the research gap identified above, this study examines whether replacing K-means clustering in the original BPA approach with a representation learning mechanism and latent space advice retrieval can improve advice reuse and training performance. The new framework, including representation learning and latent space advice retrieval, is called Encoder BPA.

This research hypothesises that the representation learning mechanism will capture more meaningful behavioural features than raw state clustering, and latent space advice retrieval will identify more relevant previous advice. As a result, the agent may reuse external advice more accurately and achieve improved training performance compared with the original K-means clustering BPA approach.

This research is expected to make several contributions. First, it proposes the Encoder BPA approach, providing an alternative advice reuse mechanism for persistent advising. Second, this research contributes by implementing persistent advising and BPA in continuous action space environments. Third, this research presents an experimental study of the proposed BPA encoder approach in partially observable reinforcement learning environments.

\section{Related Work}

\subsection{Interactive Reinforcement Learning}

Interactive Reinforcement Learning (IntRL) extends the traditional RL framework by allowing an external trainer to provide advice during the learning process~\cite{cruzSurveyInteractiveReinforcement2021}. The trainer may be a human expert or another agent with prior knowledge of the task, which can guide the learner when assistance is available~\cite{najarReinforcementLearningHuman2021,bignoldEvaluationMethodologyInteractive2021}. The difference between traditional RL and IntRL is that in IntRL, advice can influence action selection, guide exploration, and may consequently accelerate learning.

Deep Interactive Reinforcement Learning (Deep IntRL) emerged from the integration of IntRL with Deep RL, allowing agents to learn from both environmental rewards and external feedback in high-dimensional or continuous state spaces~\cite{moreiraDeepReinforcementLearning2020,nguyenBroadPersistentAdvisingApproach2023}. However, most existing Deep IntRL methods remain limited in that feedback is typically given only for the current state and is discarded after a single use. To address these limitations, recent Deep IntRL research has focused on making advice more reusable rather than treating it as one-off feedback~\cite{bignoldPersistentRulebasedInteractive2023,nguyenBroadPersistentAdvisingApproach2023}.

\subsection{Persistent Advising}

Persistent advising extends Deep IntRL by allowing the agent to retain advice from a trainer and reuse it in later decision-making~\cite{bignoldPersistentRulebasedInteractive2023}. This improves the value of each interaction and reduces the need for the trainer to provide the same guidance repeatedly.

Persistent advising includes the Probabilistic Policy Reuse (PPR) mechanism, in which the agent gradually reduces its dependence on previous advice and increases its trust in newly acquired information. In this framework, the agent can learn from the newer trainer guidance through the training process~\cite{bignoldPersistentRulebasedInteractive2023}.

State-based advice is a form of persistent advising in which the agent directly stores (state, action) pairs in a persistent memory after the trainer provides advice. When the same state is encountered again, the agent can reuse the stored advice instead of requesting new input~\cite{bignoldPersistentRulebasedInteractive2023}. This method is effective in discrete environments, where exact states are more likely to reappear. However, its main limitation is that states are not effectively generalised, which makes the method less suitable for large or continuous state spaces.

\begin{figure}[H]
    \centering
    \includegraphics[width=0.5\textwidth]{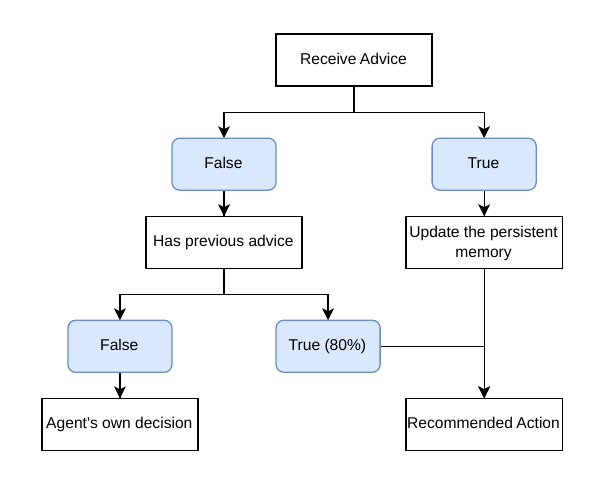}
    \caption{The PPR mechanism. When external advice is available, the learning agent applies the advised action and stores the (state, action) pair in the persistent memory. When no external advice is available, the associated action may be reused if a matching state is found in the persistent memory.}
    \label{fig:Persistent_flowchart_simple}
\end{figure}

Figure~\ref{fig:Persistent_flowchart_simple} illustrates how the state-based persistent memory is used in the learning process of Deep IntRL. In each time step, if there is external advice, the learning agent applies the advised action and stores the (state, action) pair in the persistent memory. When no external advice is available, the associated action may be reused if a matching state is found in the persistent memory.

\subsection{Broad-Persistent Advising}

Broad-Persistent Advising (BPA) builds on persistent advising by enabling advice to be generalised and reused across similar states during later decision-making~\cite{nguyenBroadPersistentAdvisingApproach2023}. This approach has two core components: persistent advising, which preserves advice for later reuse, and broad advising, which extends advice reuse from a repeated state to similar states.

In broad advising, BPA uses a clustering model to map a raw state to a cluster ID and stores advice as a (cluster\_id, action) pair rather than a (state, action) pair~\cite{nguyenBroadPersistentAdvisingApproach2023,murphyProbabilisticMachineLearning2022}. This allows advice given in one state to be reused in other states that share similar features. As a result, broad advising increases the utilisation of persistent memory, making it more suitable for large and continuous state spaces. 

In the original BPA approach, K-means clustering is used for generalisation with the elbow method, which helps determine the value of $K$. In Cart-Pole, the state consists of four continuous values. In contrast, the Domestic Robot environment uses a $64 \times 64 \times 3$ state representation, resulting in a high-dimensional input space, while the elbow method selects only four clusters. In this research, we develop a framework that uses a neural network encoder to replace the K-means algorithm in broad advising, where the learned latent representation is used for advice comparison~\cite{nguyenBroadPersistentAdvisingApproach2023}.

\subsection{Representation Learning}

Representation learning allows a model to extract useful features from raw data and express them in more compact feature representations~\cite{bengioRepresentationLearningReview2012}. Prior research generally suggests that effective representations can extract information from complex inputs and map them into a compact feature space that is better suited for downstream tasks, thereby improving learning efficiency~\cite{srinivasCURLContrastiveUnsupervised2020,stookeDecouplingRepresentationLearning2021,yaratsReinforcementLearningPrototypical2021}. In reinforcement learning, representation learning can be used to address the limitations of raw state observations. In formal terms, given an input $x \in \mathcal{X}$, a representation model learns a mapping $f_{\theta}: \mathcal{X} \rightarrow \mathcal{Z}$, where the latent representation is defined as $z = f_{\theta}(x)$~\cite{echchahedSurveyStateRepresentation2025}. A common way to compare two learned representations is cosine similarity. Given two sequence representations $z_i$ and $z_j$, it is defined as:
\begin{equation}
    \mathrm{cos}(z_i, z_j) = \frac{z_i^\top z_j}{\|z_i\|\,\|z_j\|}.
\end{equation}
Cosine similarity compares the direction of two fixed-dimensional embedding vectors rather than their magnitudes. This makes it suitable for measuring similarity of encoded state histories~\cite{chenSimpleFrameworkContrastive2020}.

In this research, representation learning is used to replace the clustering-based state abstraction adopted in the original BPA. The proposed framework trains a latent encoder using state transition data. A persistent memory cell is then stored as a pair of latent and action ($z_i$, action), where $z_i$ represents the encoded state history. During advice retrieval, the current latent representation can be compared with stored representations using cosine similarity.

\section{Methodology}

The BPA method enables advice to be reused across similar states. However, the broad advising component relies on K-means clustering, which may be limited in high-dimensional environments. In the methodology section, we describe the Encoder BPA framework, which may enhance the original BPA approach by using representation learning and latent space advice retrieval.

\subsection{Encoder Pre-training}

In representation learning, we use a supervised learning method to train an encoder network to capture action-discriminative features. In this process, a teacher network continuously interacts with the environment to collect transition data. The collected data are then used to train an encoder together with a prediction head. At each time step $t$, the input to the encoder is a sequence of the most recent $L$ states, where the state-history length is fixed at $L = 16$ across all experimental environments:
\begin{equation}
    h_t = (s_{t-L+1}, s_{t-L+2}, \dots, s_t),
\end{equation}
and the corresponding training target is the action $a_t$. The encoder $f_{\theta}$ maps the input sequence $h_t$ to a latent representation $z_t$, defined as:
\begin{equation}
    z_t = f_{\theta}(h_t).
\end{equation}
A prediction head, implemented as a multilayer perceptron (MLP), takes $z_t$ as input and predicts the action $\hat{a}_t$. The encoder and prediction head are jointly trained using cross-entropy loss for discrete actions or mean squared error loss for continuous actions.

After the encoder pre-training stage, the prediction head is discarded, and the encoder parameters are frozen. The pre-trained encoder is then used to generate latent representations for downstream use in the advising framework.

\subsection{Encoder BPA}

The main differences between the original BPA approach and the Encoder BPA approach are in the broad advising part. After the pre-training, the Encoder BPA approach maintains a fixed-length queue containing the most recent $L$ states, $s_{t-L+1}, \ldots, s_t$ at each time step and uses the pre-trained encoder to obtain a latent representation $z_i$ from the state history.

In the proposed framework, we retain the original persistent advising structure, which uses the PPR strategy to enable probabilistic advice reuse. Our framework also uses the original initial reuse probability of 80\%. However, it adopts a lower decay rate in our Encoder BPA framework so that advice stored in persistent memory can be reused for a longer period. When external advice is available, the learning agent applies the advised action and stores the ($z_i$, action) pair in the persistent memory. When no external advice is available, the associated action may be reused in latent space advice retrieval if a similar $z_i$ is found in the persistent memory.

\begin{figure}[H]
    \centering
    \includegraphics[width=0.5\textwidth]{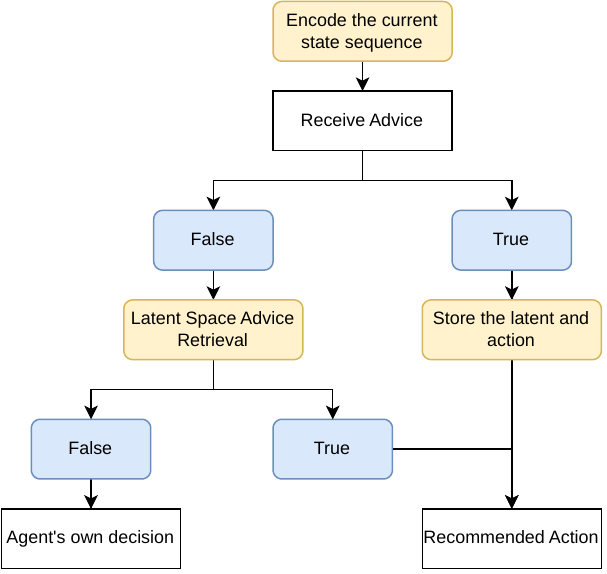}
    \caption{The PPR mechanism in Encoder BPA. In Encoder BPA, a trained encoder is used to map a state sequence into a latent representation at every time step. When external advice is available, the learning agent applies the advised action and stores the (latent, action) pair in the persistent memory. When no external advice is available, the associated action may be reused in latent space advice retrieval.}
    \label{fig:advice_retrieval}
\end{figure}

\subsection{Latent Space Advice Retrieval}

In Encoder BPA, advice retrieval includes three components: the similarity threshold $\tau$, the top-(k) aggregation strategy, and the consistency threshold $\delta$. Our framework first measures similarity in latent space between the current state-sequence representation and stored memories. Specifically, for each latent representation $z_i$ at time step $t$, we filter out all similar previous latents using a cosine similarity threshold condition as follows:

\begin{equation}
    \operatorname{sim}(z_c, z_i) > \tau
\end{equation}

Both $z_c$ and $z_i$ are latent representations of state sequences with the same length. If the similarity between them exceeds the threshold $\tau$, the current situation is considered sufficiently similar to a previously advised situation, and the corresponding stored action can be reused. 

Then, the top-(k) aggregation is used to select similar previous latents from the memory set, and a majority vote or averaging strategy is used to form a candidate action~\cite{mestoConsensusParadoxWhen2026}. The corresponding reuse probability can also be calculated by averaging the reuse rates of the selected memories.

To ensure that the retrieved action can be reused reliably, this study uses a consistency threshold $\delta$ to evaluate the action consistency of the memory set. In a discrete action space, majority voting is applied, and the consistency threshold specifies the minimum proportion of the majority action in top-(k) similar memories. In a continuous action space, the retrieved actions are averaged to obtain the candidate action. Their consistency is evaluated using the mean action distance, and the candidate action is reused only when the mean action distance is below the consistency threshold.

\subsection{Experimental Design}

In this research, the average reward over the most recent 100 episodes across 9 different random seeds is used to evaluate the performance of different IntRL approaches. A higher average reward indicates that the agent achieves better training performance. Another important metric is the number of advice reuses. This value increases only when the persistent memory successfully provides a piece of advice to reuse. Therefore, it directly reflects how often the persistent memory contributes to the action selection during training.

For model comparison, Deep Q-Network (DQN) is used as the reinforcement learning backbone in discrete action spaces, while Soft Actor-Critic (SAC) is used in continuous action spaces. Since the original BPA method relies on an $\epsilon$-greedy strategy, this study follows the original BPA implementation for the DQN setting. In continuous action spaces, the PPR mechanism in BPA behaves similarly to persistent advising, allowing a fair comparison.

\subsection{Experimental Environments}

\begin{figure}[t]
    \centering
    \begin{subfigure}[t]{0.3\linewidth}
        \centering
        \includegraphics[width=\linewidth]{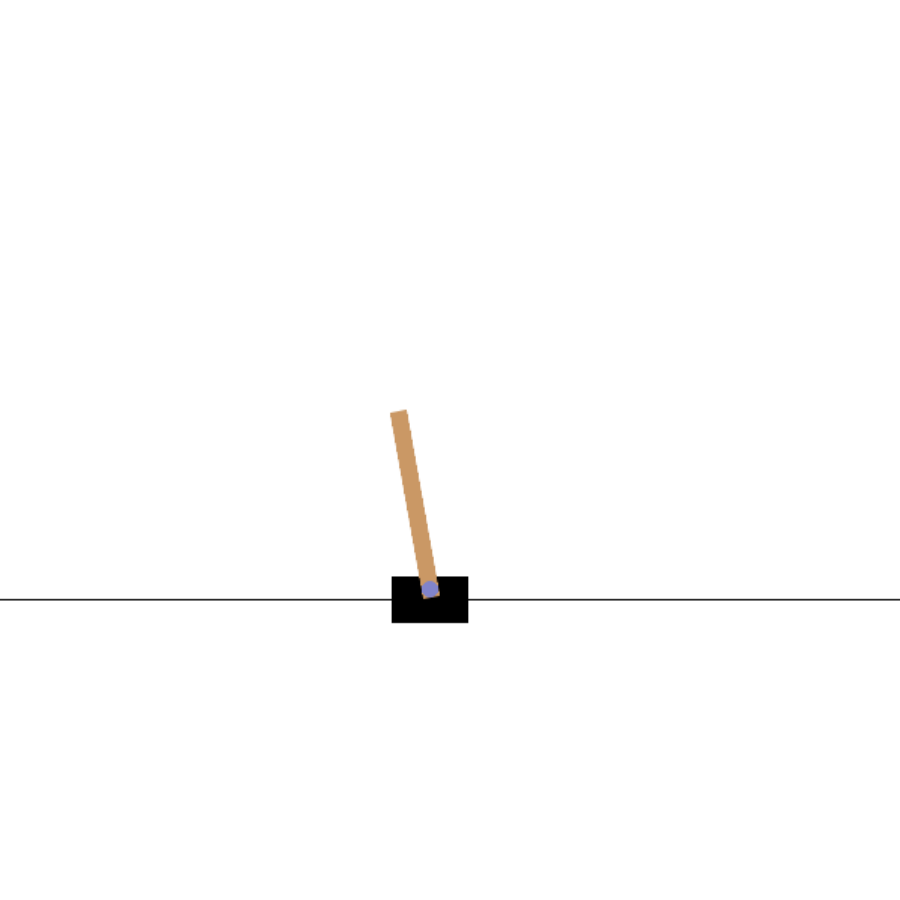}
        \caption{Cart-Pole}
        \label{fig:cartpole_env}
    \end{subfigure}
    \hfill
    \begin{subfigure}[t]{0.3\linewidth}
        \centering
        \includegraphics[width=\linewidth]{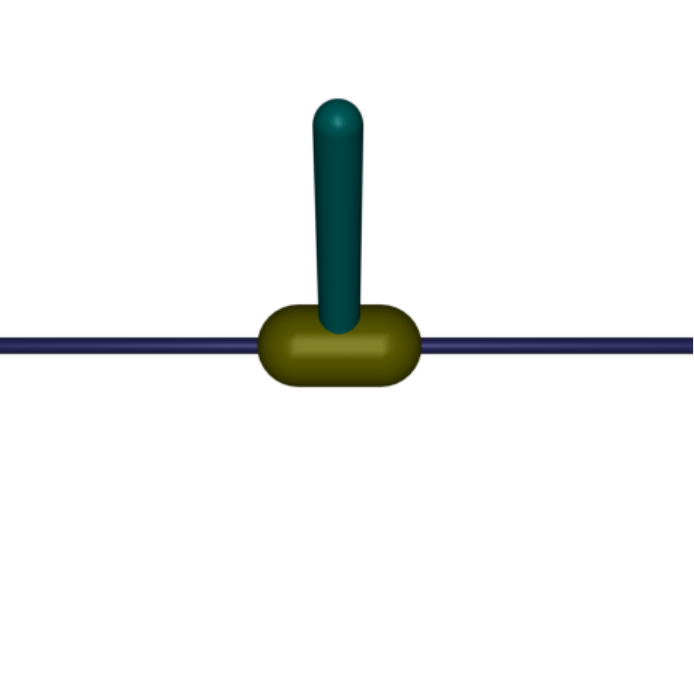}
        \caption{Inverted Pendulum}
        \label{fig:invertedpendulum_env}
    \end{subfigure}
    \hfill
    \begin{subfigure}[t]{0.3\linewidth}
        \centering
        \includegraphics[width=\linewidth]{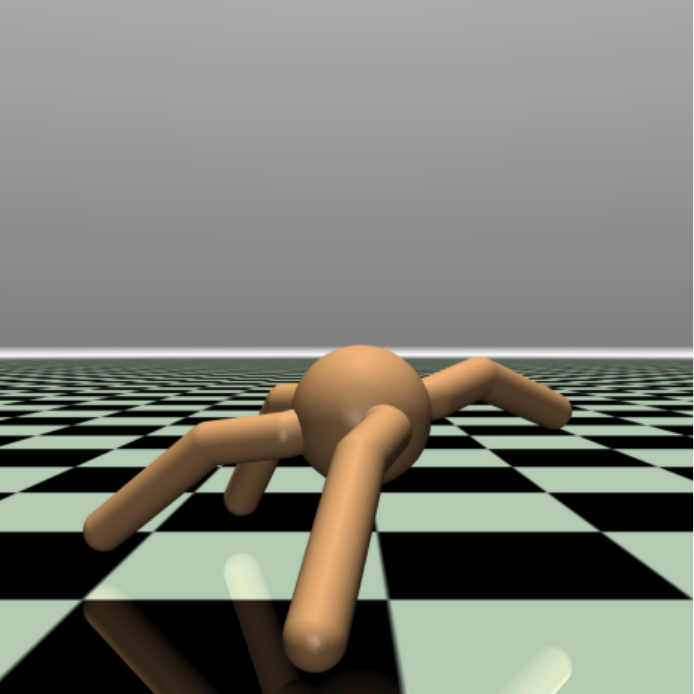}
        \caption{Ant}
        \label{fig:ant_env}
    \end{subfigure}

    \caption{The environments used in the experiments.}
    \label{fig:environments}
\end{figure}

\subsubsection{Cart-Pole}

Cart-Pole is a classic reinforcement learning control environment that aims to balance a pole by controlling a cart under the pole using forces in the left or right directions~\cite{brockmanOpenAIGym2016}. The state is continuous and in 4 dimensions, and the action is discrete, representing the direction of the force. The agent can continuously obtain rewards by balancing the pole at each time step.

\subsubsection{Inverted Pendulum}

The second experiment is in the Inverted Pendulum environment. This environment is similar to Cart-Pole in terms of rules, but its action space is continuous. The action space is one-dimensional and ranges from $-3.0$ to $3.0$, where negative and positive values represent forces applied in opposite directions.

\subsubsection{MuJoCo Ant}

The third experiment is in the MuJoCo Ant environment. In this environment, an agent learns to coordinate a four-legged robot to move forward efficiently while maintaining balance. The state is continuous and in 105 dimensions, including position, orientation, and joint information. The action is continuous and in 8 dimensions, controlling the torques applied to the legs of the agent. The reward encourages the ant to move forward efficiently while maintaining balance.

\subsubsection{POMDP Ant}

The fourth experiment is conducted in a partially observable Ant environment. Unlike the original Ant environment, under the POMDP setting, the agent cannot observe the complete environment state and must make decisions based only on partial observations. To construct the POMDP settings, the state variables at indices 13 and 14 are masked and therefore unavailable to the learning agent. Indices 13 to 14 represent the x, y coordinate velocity of the torso. In addition, the agent of a POMDP environment learns from a sequence of recent observations rather than relying only on the current state. In this experiment, the observation history consists of the four most recent observations. This is distinct from the Encoder BPA history length $L = 16$, which is used as the input sequence to the latent encoder for advice retrieval.

In this experimental environment, the teacher network is trained using interactive reinforcement learning, in which the teacher network of the original Ant environment provides external advice. After training, the external advice is removed, and the teacher network interacts with the environment independently. An evaluation curve is then generated to assess the learned policy and verify the effectiveness of the advice provided by the teacher network.

\subsection{Advisor Configuration}

The advisor is controlled by two parameters: frequency and accuracy. The frequency represents how often the advisor gives advice. The accuracy parameter controls the reliability of the advice. There are two advisor settings used in this experiment, as follows:

\begin{itemize}
    \item Advisor 1: {frequency: 0.5, accuracy: 1.0}
    \item Advisor 2: {frequency: 0.5, accuracy: 0.9}
\end{itemize}

By using these two advisor configurations, the experiments can compare both the effectiveness and robustness of different methods. In practice, the frequency of human advice varies across tasks. This study therefore uses a fixed advice frequency as a reference setting.

\section{Results and Discussion}

This chapter reports and analyses the experimental results obtained from the Cart-Pole, Inverted Pendulum, Ant, and POMDP Ant environments.

\begin{figure*}[t]
    \centering
    \begin{subfigure}[t]{0.32\textwidth}
        \centering
        \includegraphics[width=\linewidth]{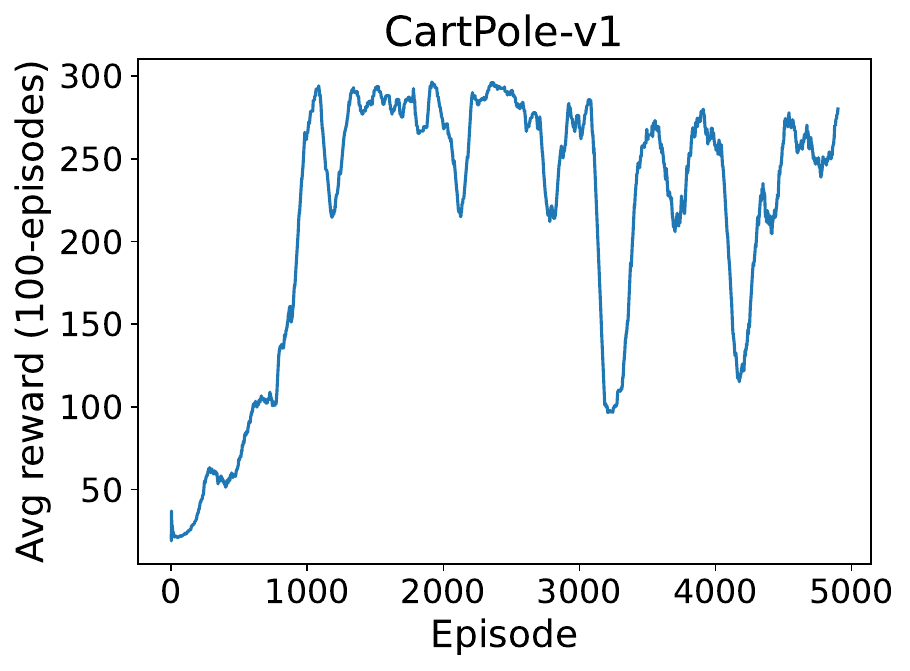}
        \caption{Teacher network in the Cart-Pole environment. The DQN teacher network is trained for 5000 episodes and evaluated by the average reward.}
        \label{fig:cartpole_teacher}
    \end{subfigure}
    \hfill
    \begin{subfigure}[t]{0.32\textwidth}
        \centering
        \includegraphics[width=\linewidth]{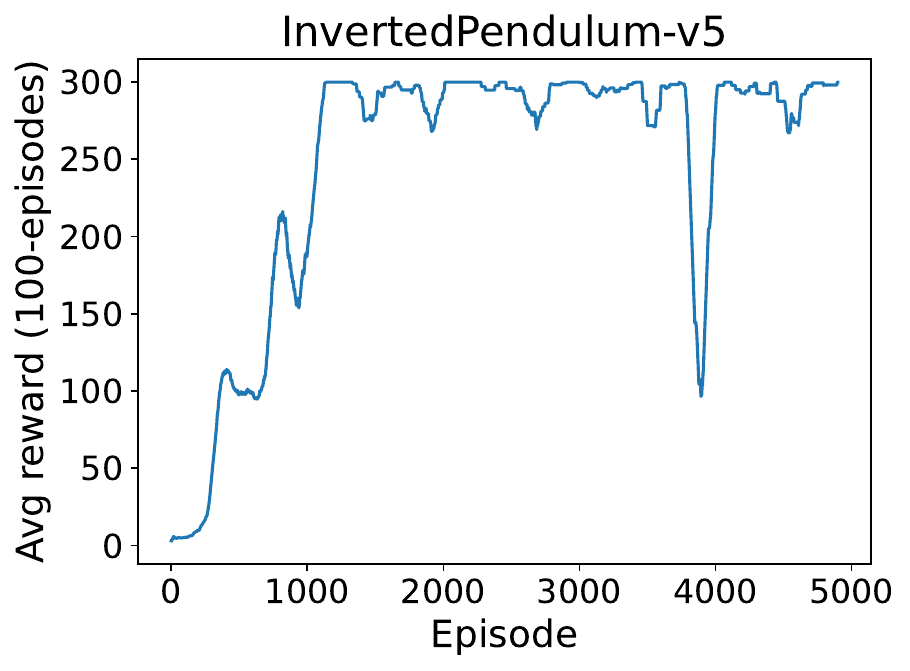}
        \caption{Teacher network in the Inverted Pendulum environment. The SAC teacher network is trained for 5000 episodes and evaluated by the average reward.}
        \label{fig:invertedpendulum_teacher}
    \end{subfigure}
    \hfill
    \begin{subfigure}[t]{0.32\textwidth}
        \centering
        \includegraphics[width=\linewidth]{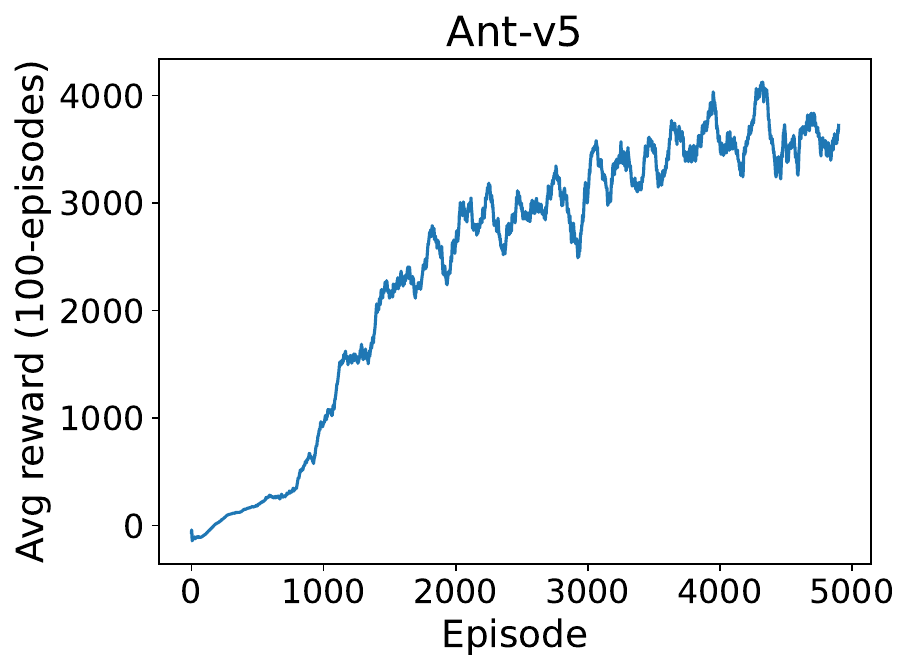}
        \caption{Teacher network in the Ant environment. The SAC teacher network is trained for 5000 episodes and evaluated by the average reward.}
        \label{fig:ant_teacher}
    \end{subfigure}

    \vspace{0.1cm}

    \begin{subfigure}[t]{0.32\textwidth}
        \centering
        \includegraphics[width=\linewidth]{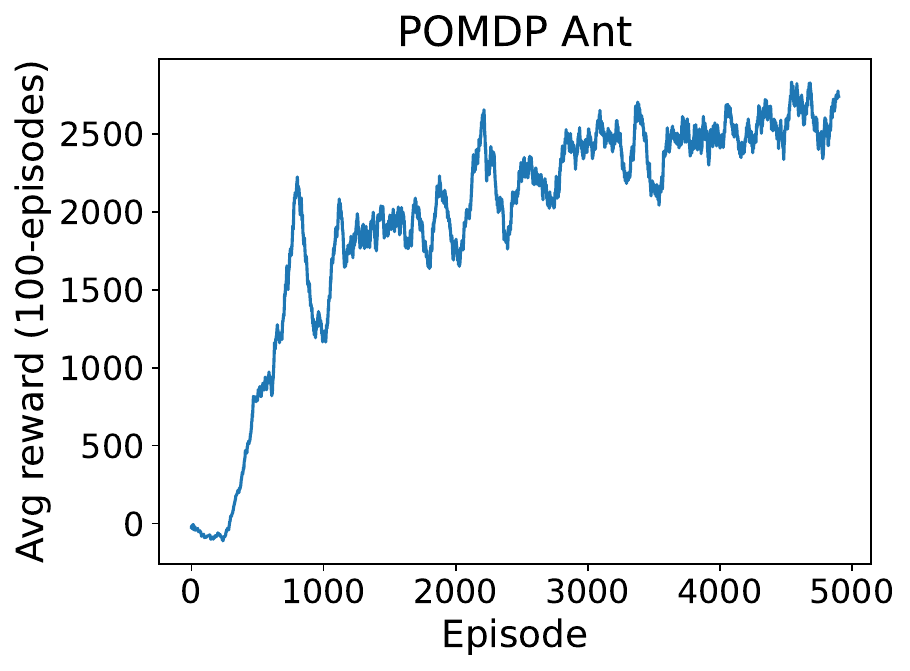}
        \caption{Teacher network in the POMDP environment. In the POMDP experiment, the teacher network of the original Ant environment provides external advice to accelerate training.}
        \label{fig:pomdp_teacher}
    \end{subfigure}
    \hspace{0.04\textwidth}
    \begin{subfigure}[t]{0.32\textwidth}
        \centering
        \includegraphics[width=\linewidth]{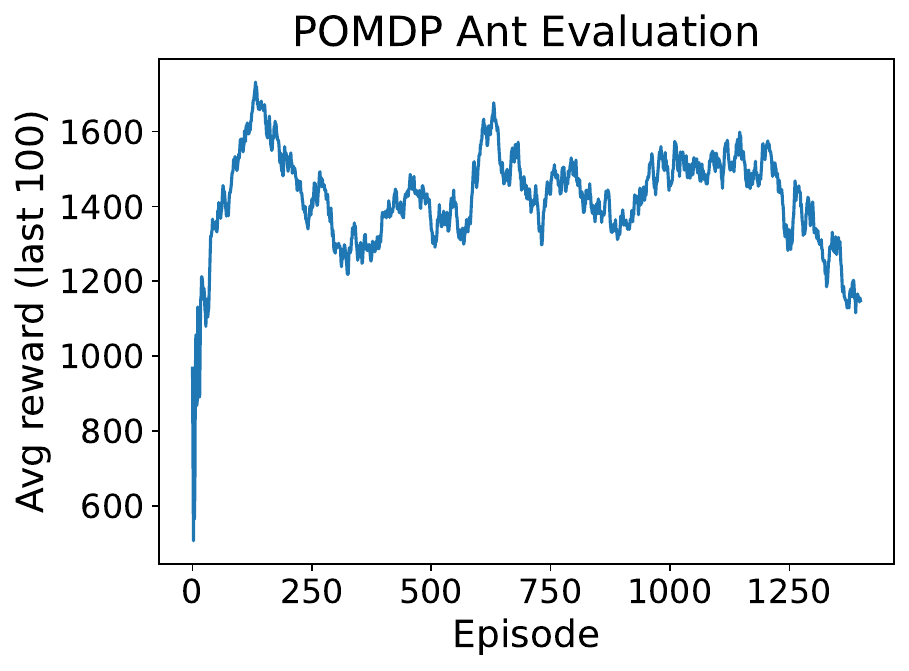}
        \caption{The evaluation curve of the teacher network in the POMDP environment. The teacher network is evaluated by using a deterministic policy with exploration disabled.}
        \label{fig:pomdp_evaluation}
    \end{subfigure}

    \caption{The training curve of the teacher network in different environments.}
    \label{fig:teacher_result}
\end{figure*}

\subsection{Teacher Network Performance}

This section presents the training performance of the teacher networks across different environments. Figure~\ref{fig:cartpole_teacher} shows that the training curve of the teacher network in the Cart-Pole environment gradually converges to an average reward of approximately 300, which represents high performance. Figure~\ref{fig:invertedpendulum_teacher} shows that the training curve gradually converges to an average reward of approximately 300 and maintains high performance after 2000 episodes. Figure~\ref{fig:ant_teacher} shows that the training reward of the teacher network in the MuJoCo Ant environment gradually increases to approximately 3000 after 1000 episodes. In addition, Figure~\ref{fig:pomdp_teacher} shows that, with guidance from the Ant teacher network, the teacher agent in the POMDP Ant environment gradually converges to an average reward of approximately 2500 per episode. Figure~\ref{fig:pomdp_evaluation} further demonstrates that when interacting with the environment independently without external advice, the trained teacher network in the POMDP Ant environment maintains an average reward of around 1400 per episode.

\subsection{K-means and Encoder}

In the original BPA approach, the elbow method is used to select the $K$ value for K-means clustering in both the Cart-Pole and the robotic environment. To maintain methodological consistency, this section presents results from the elbow method. Table~\ref{tab:kmeans_results} shows the $K$ values applied to each environment.

\begin{table}[htbp]
    \caption{The results of the elbow method.}
    \centering
    \resizebox{0.85\columnwidth}{!}{
        \begin{tabular}{|c|c|c|c|c|}
            \hline
            \textbf{Parameter}
            & \textbf{Cart-Pole}
            & \textbf{Inv. Pend.}
            & \textbf{Ant}
            & \textbf{POMDP} \\
            \hline
            $K$ Value
            & 3 & 4 & 2 & 4 \\
            \hline
        \end{tabular}
    }
    \label{tab:kmeans_results}
\end{table}

To evaluate whether the learned latent representations capture action-discriminative features associated with the teacher policy, this part presents Spearman correlation results for the LSTM and Mamba encoders across different environments. Table~\ref{tab:correlation_results} shows the Spearman correlation results in different environments.

\begin{table}[htbp]
    \caption{Spearman correlation between representation cosine similarity and action distance across different environments and encoders.}
    \centering
    \resizebox{0.85\columnwidth}{!}{
        \begin{tabular}{|c|c|c|c|c|}
            \hline
            \textbf{Encoder}
            & \textbf{Cart-Pole}
            & \textbf{Inv. Pend.}
            & \textbf{Ant}
            & \textbf{POMDP} \\
            \hline
            LSTM
            & -0.816 & -0.986 & -0.870 & -0.886 \\
            \hline
            Mamba
            & -0.819 & -0.967 & -0.867 & -0.889 \\
            \hline
        \end{tabular}
    }
    \label{tab:correlation_results}
\end{table}

The Spearman correlation results show a strong negative monotonic relationship between representation similarity and action distance. This means that when two state sequences have more similar latent representations, the distance between their corresponding teacher actions tends to be smaller. Therefore, the results support the reliability of using a similarity threshold to identify relevant previous advice for action selection.

\begin{figure*}[t]
    \centering

    \begin{subfigure}{0.47\textwidth}
        \centering
        \includegraphics[width=\linewidth]{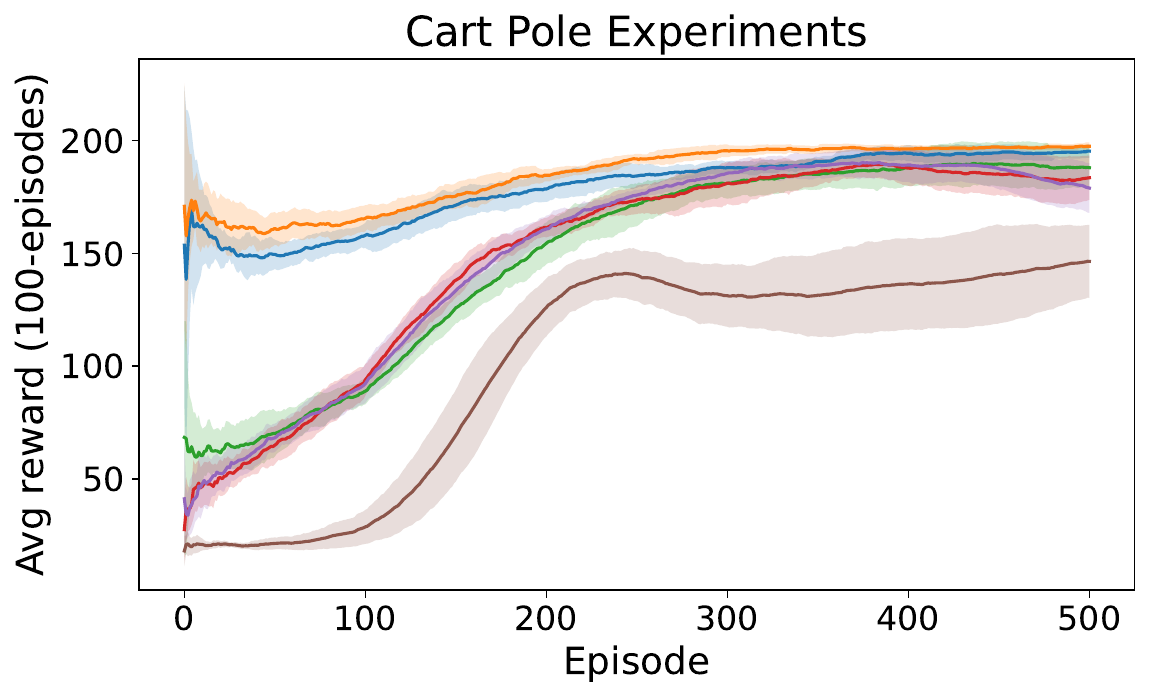}
        \caption{The Cart-Pole Experiments. This visualisation evaluates the performance of different IntRL approaches in the discrete action environment.}
        \label{fig:cartpole}
    \end{subfigure}
    \hfill
    \begin{subfigure}{0.47\textwidth}
        \centering
        \includegraphics[width=\linewidth]{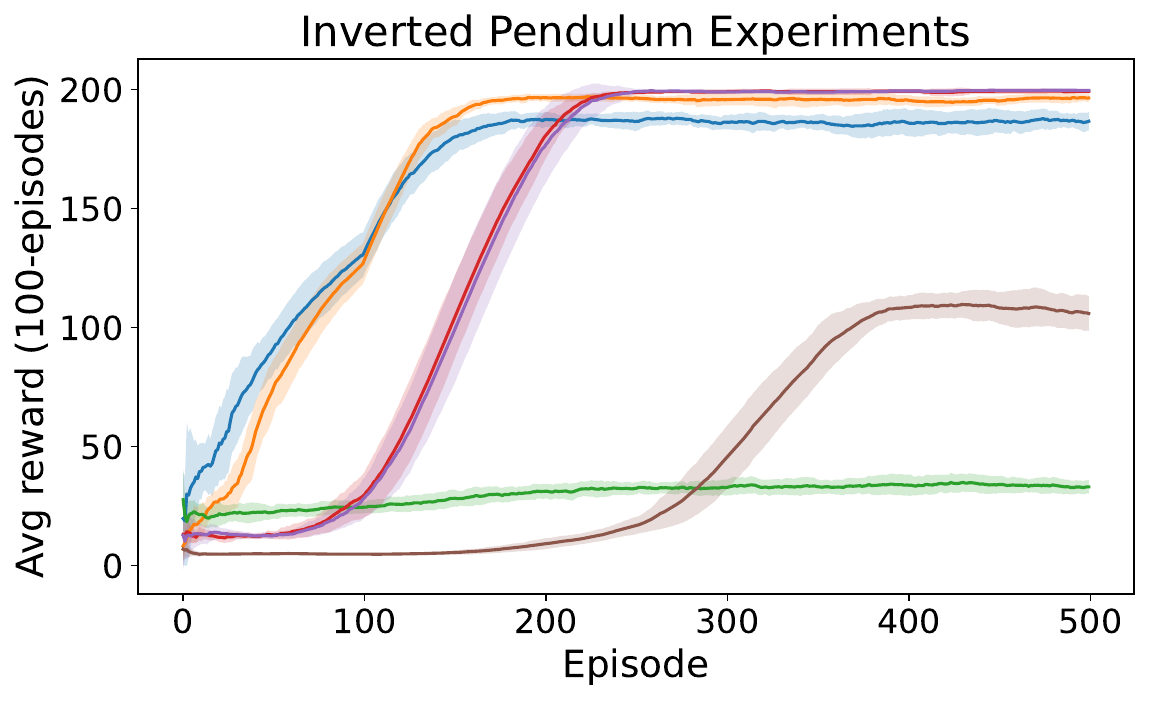}
        \caption{The Inverted Pendulum experiments. This visualisation evaluates the performance of different IntRL approaches in the continuous action environment.}
        \label{fig:invertedpendulum}
    \end{subfigure}

    \vspace{0.1cm}

    \begin{subfigure}{0.47\textwidth}
        \centering
        \includegraphics[width=\linewidth]{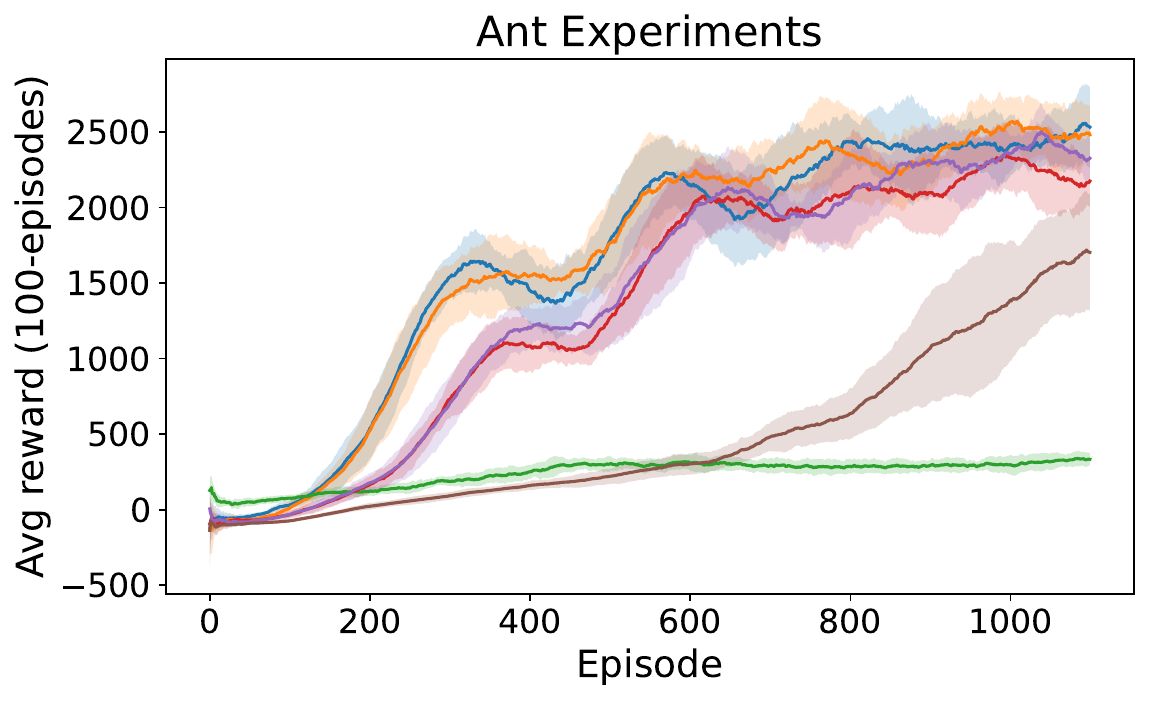}
        \caption{The Ant experiments. This visualisation evaluates the performance in the complex control environment.}
        \label{fig:ant}
    \end{subfigure}
    \hfill
    \begin{subfigure}{0.47\textwidth}
        \centering
        \includegraphics[width=\linewidth]{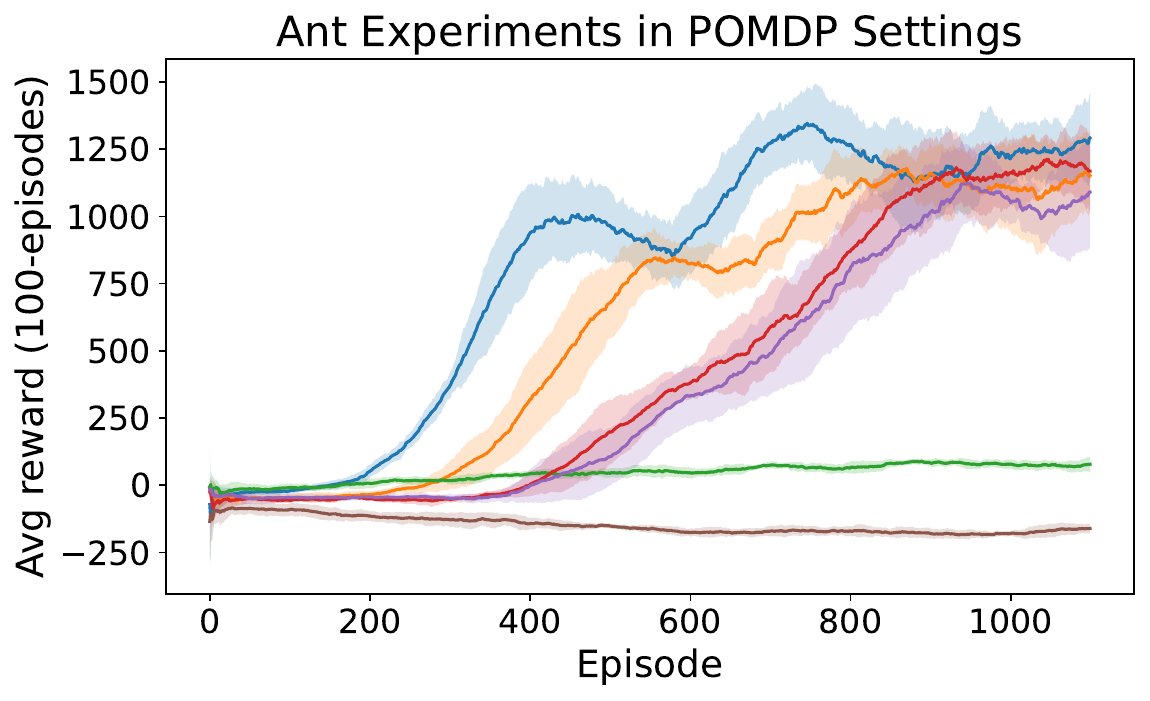}
        \caption{The POMDP Ant experiments. This visualisation evaluates the performance in the POMDP environment.}
        \label{fig:pomdp}
    \end{subfigure}

    \vspace{0.1cm}
    
    \begin{subfigure}{\textwidth}
        \centering
        \includegraphics[width=\linewidth]{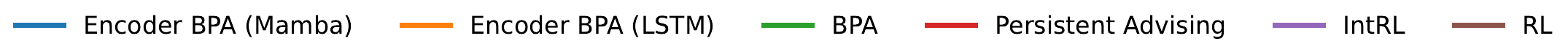}
        \label{fig:legend}
    \end{subfigure}

    \caption{Comparison of different interactive reinforcement learning approaches across different environments. The solid lines represent the average reward across multiple random seeds, while the shaded regions represent the corresponding standard deviation around the mean.}
    \label{fig:experiments}
\end{figure*}

\subsection{Cart-Pole}

The Cart-Pole experiments evaluate whether the proposed Encoder BPA approach improves training performance by reusing stored advice in a discrete action environment. Based on the methodology, latent space advice retrieval includes a cosine similarity threshold, a top-(k) aggregation, and a consistency threshold. The main settings are as follows:

\begin{itemize}
    \item Similarity threshold $\tau$: 0.95
    \item Top-(k) aggregation: top-5 majority voting
    \item Consistency threshold $\delta$: 0.5
\end{itemize}

For a discrete action space, the consistency threshold represents the minimum proportion of the majority action. Since Cart-Pole has only two actions, 0 and 1, the majority action must appear in at least 3 of the top-5 similar memories.

The training curve in Figure~\ref{fig:cartpole} shows that the Encoder BPA approach using an LSTM or Mamba encoder achieves the best early training performance. These results suggest that representation learning and latent space advice retrieval can effectively support the reuse of previous advice in a discrete action space.

\subsection{Inverted Pendulum}

The Inverted Pendulum experiments evaluate whether the proposed persistent advising approach improves training performance in a continuous-action environment. The advice retrieved in the Inverted Pendulum task is a continuous value. Therefore, the averaging strategy is used to select the candidate action.

\begin{itemize}
    \item Similarity threshold $\tau$: 0.995
    \item Top-(k) aggregation: top-5 averaging strategy
    \item Consistency threshold $\delta$: 0.5
\end{itemize}

In a continuous action space, the consistency threshold is evaluated using the mean action distance.

As shown in Figure~\ref{fig:invertedpendulum}, the training performance of the Encoder BPA increases in the early training stage. However, in later episodes, the Encoder BPA agent may reuse inaccurate or inappropriate advice through the averaging strategy, which can negatively affect its action selection and consequently reduce the rewards it receives.

\subsection{MuJoCo Ant}

The Ant experiments test the proposed approach in a more complex control environment. At each step, the agent must use information about body position, joint configuration, and velocity to coordinate several joints simultaneously.

\begin{itemize}
    \item Similarity threshold $\tau$: 0.97
    \item Top-(k) aggregation: top-5 averaging strategy
    \item Consistency threshold $\delta$: 0.5
\end{itemize}

Since the Ant environment uses continuous state and action spaces, the averaging strategy and mean action distance are applied.

Figure~\ref{fig:ant} shows that the Encoder BPA approach achieves a greater performance improvement during the early stages of training than the other IntRL approaches in the Ant environment. This suggests that the proposed approach is effective in complex control environments with high-dimensional state and continuous action spaces.

\subsection{POMDP Ant}

The POMDP experiments evaluate whether the proposed approach remains effective when the agent cannot directly observe the complete state of the environment.

\begin{itemize}
    \item Similarity threshold $\tau$: 0.97
    \item Top-(k) aggregation: top-5 averaging strategy
    \item Consistency threshold $\delta$: 0.5
\end{itemize}

The POMDP experiment uses the same advice retrieval configuration as the MuJoCo Ant environment.

Figure~\ref{fig:pomdp} shows that the Encoder BPA approach substantially improved early-stage training performance in the partially observable environment. By encoding the sequence of recent states into a latent representation, the encoder may capture useful temporal information that may not be fully available from a single observation or a short observation sequence. This may enable the Encoder BPA agent to retrieve relevant previous advice, resulting in better performance during the early stages of training.

\subsection{Advice Reuse Frequency}

This section compares the frequency of advice reuse in the persistent memory between different approaches, including state-based persistent advising, BPA, and Encoder BPA.

\begin{table}[htbp]
    \caption{Average number of advice reuses across multiple experiments.}
    \centering
    \resizebox{0.85\columnwidth}{!}{%
        \begin{tabular}{|c|c|c|c|c|}
            \hline
            \textbf{}
            & \textbf{Cart-Pole}
            & \textbf{Inv. Pend.}
            & \textbf{Ant}
            & \textbf{POMDP} \\
            \hline
            Persistent
            & 18 & 1 & 0 & 0 \\
            \hline
            BPA
            & 6032 & 5892 & 141952 & 73990 \\
            \hline
            LSTM
            & 28025 & 16553 & 119721 & 63288 \\
            \hline
            Mamba
            & 27737 & 17813 & 127132 & 99505 \\
            \hline
        \end{tabular}
    }
    \label{tab:reuse_results}
\end{table}

Table~\ref{tab:reuse_results} shows that the Encoder BPA provides more opportunities to reuse previous advice in similar states compared to state-based persistent advising and the original BPA approach in most cases. In continuous state spaces, exact state revisits are difficult within a limited number of steps, resulting in an extremely low average reuse frequency for persistent advising.

\begin{figure*}[t]
    \centering
    \begin{subfigure}[t]{0.32\textwidth}
        \centering
        \includegraphics[width=\linewidth]{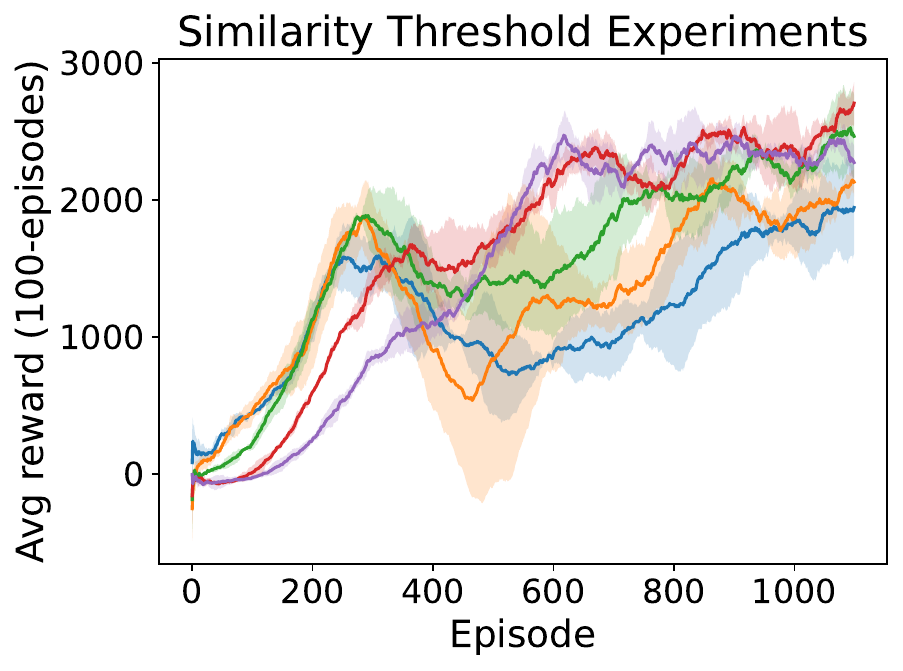}
        \vspace{0.05cm}
        \includegraphics[width=0.8\linewidth]{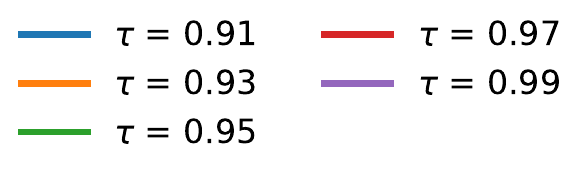}
        \caption{Different similarity thresholds in the Ant environment. In latent space advice retrieval, the top-5 averaging strategy and the 0.5 consistency threshold remain the same.}
        \label{fig:similarity_threshold}
    \end{subfigure}
    \hfill
    \begin{subfigure}[t]{0.32\textwidth}
        \centering
        \includegraphics[width=\linewidth]{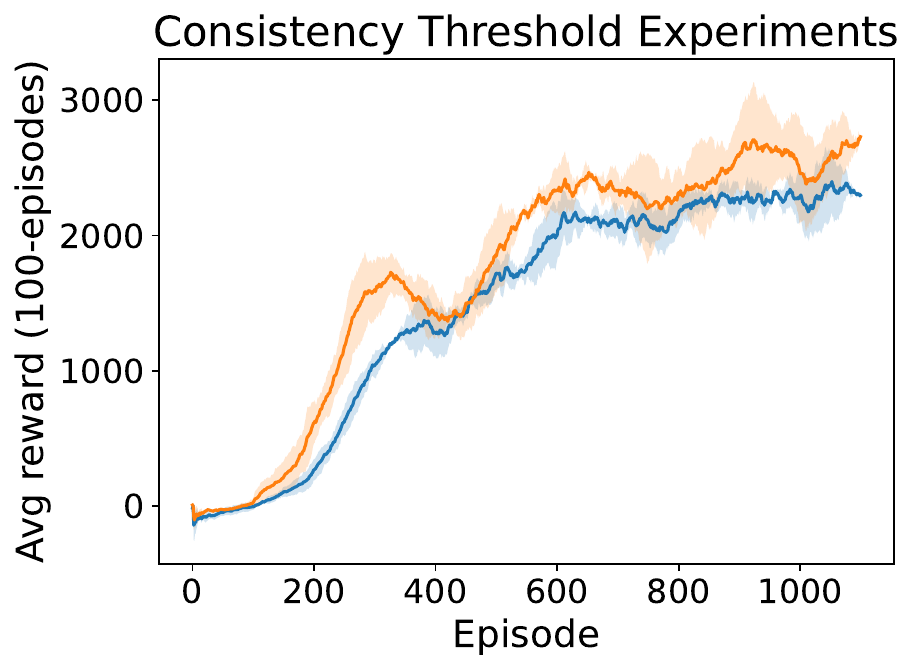}
        \vspace{0.05cm}
        \includegraphics[width=0.7\linewidth]{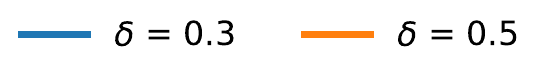}
        \caption{Different consistency thresholds in the Ant environment. In latent space advice retrieval, the 0.97 similarity threshold and the top-5 averaging strategy remain the same.}
        \label{fig:consistency_threshold}
    \end{subfigure}
    \hfill
    \begin{subfigure}[t]{0.32\textwidth}
        \centering
        \includegraphics[width=\linewidth]{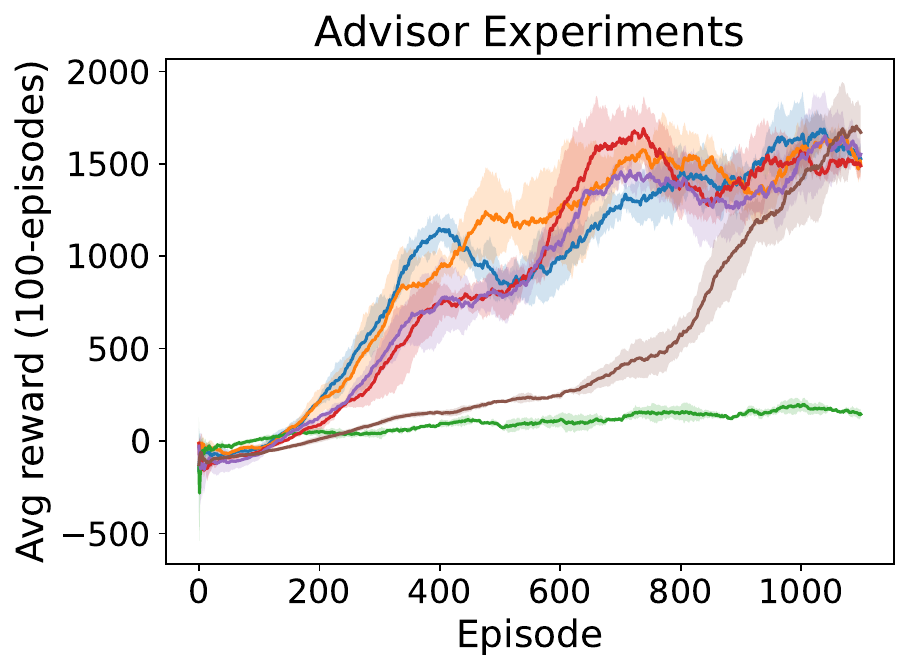}
        \vspace{0.05cm}
        \includegraphics[width=0.8\linewidth]{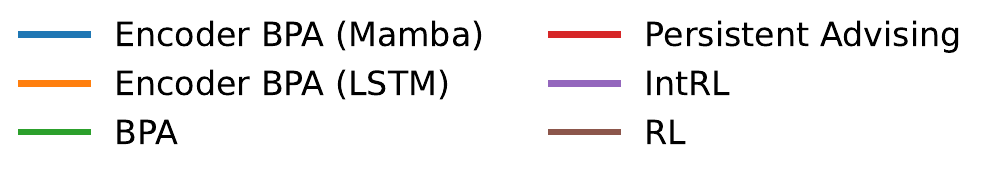}
        \caption{Different advisor in the Ant environment. The advisor is Advisor 2, as defined in the Experimental Design section, with an accuracy of 90\%. The advice retrieval settings remain unchanged.}
        \label{fig:advisor}
    \end{subfigure}
    
    \caption{Supporting experiments in the MuJoCo Ant environment.}
    \label{fig:supporting_experiments}
\end{figure*}

\subsection{Supporting Experiments}

This section provides supporting experiments to examine how different advisor configurations and threshold values affect the agent's learning performance. This section uses the average reward across 9 different random seeds. The experimental results for different similarity thresholds are shown in Figure~\ref{fig:similarity_threshold}. The results show that a lower similarity threshold provides more opportunities to reuse previous advice, leading to better performance during the early stages of training. However, in later episodes, a low threshold and the averaging strategy continue to incorporate advice that is no longer appropriate, ultimately preventing the agent from converging. 

Secondly, the results obtained with different consistency thresholds are shown in Figure~\ref{fig:consistency_threshold}. Reducing the consistency threshold from 0.5 to 0.3 substantially decreased the number of advice reuses, suggesting that many candidate actions were rejected under the stricter threshold. Based on the training curve, 0.5 achieved better training performance and was therefore selected for the Ant experiments. 

In addition, the experimental results obtained using another advisor whose accuracy was reduced to 0.9 are shown in Figure~\ref{fig:advisor}. The results show that the Encoder BPA approach remains sensitive to noisy advice, similar to other IntRL approaches. The encoder only enables previous advice to be retrieved and reused for similar states, but it does not directly evaluate or correct the quality of the advice.

\subsection{Limitations}

Some limitations of the current Encoder BPA approach are discussed below. Even when different encoders use the same network architecture, training procedure, and hyperparameter settings, their performance may still vary, resulting in differences in the quality of advice retrieval.

Second, the latent space advice retrieval mechanism still needs further improvement. Although the method can accelerate training in the early stage, its performance may decrease later in continuous action environments. In addition, the performance of Encoder BPA is sensitive to the similarity and consistency thresholds. A small change in either threshold can cause a large change in the frequency of advice reuse. Therefore, more experiments may be needed to find suitable threshold values for each environment.

Finally, Encoder BPA requires additional computation at each training step. Before retrieving previous advice, the encoder must process the current state sequence and generate its latent representation. Nevertheless, this computational cost may be acceptable in human-in-the-loop reinforcement learning scenarios where human trainers are unable to provide advice frequently.

\subsection{Future Direction}

Future work could explore contrastive learning to improve the stability of encoder training. The loss function could be designed to place state sequences associated with similar actions closer together in the latent space, while increasing the distance between latents associated with dissimilar actions~\cite{khoslaSupervisedContrastiveLearning2020,chuangDebiasedContrastiveLearning2020}. This may produce a more structured latent space, improve advice retrieval across different types of encoders and training runs, and may reduce the sensitivity of the similarity and consistency thresholds.

The top-(k) aggregation in latent space advice retrieval can also be adjusted during agent training. Although combining multiple retrieved memories is useful in the early stage, mismatched advice may negatively affect the candidate action in a continuous action environment. Future work could gradually reduce the value of (k), or increase the decay rate of the persistent memory during training. These changes would reduce the influence of mismatched memories and allow the agent to rely more on its own policy as training progresses~\cite{nguyenBroadPersistentAdvisingApproach2023}.

\section{Conclusion}

The experimental results show that Encoder BPA improves early-stage training performance. This improvement may come partly from the additional teacher-action supervision and partly from the learned representations for advice retrieval. The Cart-Pole, Inverted Pendulum, and Ant experiments showed that the proposed approach could improve early-stage learning across discrete action, continuous action, and complex control tasks. The greater improvement in POMDP Ant further suggests that the state-history encoding of the proposed BPA approach can make better use of teacher advice when the current observation does not contain complete state information. Nevertheless, the results also show that performance could be negatively affected when the averaging strategy adopts inappropriate advice in a continuous action space.

Future experiments can focus on applying contrastive learning to the Encoder pre-training process so that state sequences linked to similar teacher actions are represented closer together. Additionally, more experiments can be conducted using dynamic advice retrieval conditions. The agent can rely more on stored advice at the beginning and gradually depend more on its own policy.

\section*{Acknowledgments}

The experiments reported in this work were performed on the UNSW compute cluster Katana
(DOI: 10.26190/669X-A286).

\balance
\bibliography{references}

@article{millanHumanFeedbackContinuous2019,
	title = {Human feedback in continuous actor-critic reinforcement learning},
	language = {en},
	journal = {27th European Symposium on Artificial Neural Networks, Computational Intelligence and Machine Learning},
	author = {Millan, Cristian and Fernandes, Bruno and Cruz, Francisco},
	year = {2019},
}

@inproceedings{mestoConsensusParadoxWhen2026,
	address = {Singapore},
	title = {The {Consensus} {Paradox}: {When} {Low} {Disagreement} {Leads} to {Catastrophic} {Failure} in {Multi}-teacher {Reinforcement} {Learning}},
	isbn = {9789819549726},
	shorttitle = {The {Consensus} {Paradox}},
	doi = {10.1007/978-981-95-4972-6_33},
	language = {en},
	booktitle = {{AI} 2025: {Advances} in {Artificial} {Intelligence}},
	publisher = {Springer Nature},
	author = {Mesto, Maher and Cruz, Francisco},
	editor = {Liu, Miaomiao and Yu, Xin and Xu, Chang and Song, Yiliao},
	year = {2026},
	pages = {426--438},
}

@article{bignoldPersistentRulebasedInteractive2023,
	title = {Persistent rule-based interactive reinforcement learning},
	volume = {35},
	issn = {1433-3058},
	url = {https://doi.org/10.1007/s00521-021-06466-w},
	doi = {10.1007/s00521-021-06466-w},
	language = {en},
	number = {32},
	urldate = {2026-03-09},
	journal = {Neural Computing and Applications},
	author = {Bignold, Adam and Cruz, Francisco and Dazeley, Richard and Vamplew, Peter and Foale, Cameron},
	month = nov,
	year = {2023},
	pages = {23411--23428},
}

@misc{brockmanOpenAIGym2016,
	title = {{OpenAI} {Gym}},
	url = {http://arxiv.org/abs/1606.01540},
	doi = {10.48550/arXiv.1606.01540},
	urldate = {2026-08-08},
	publisher = {arXiv},
	author = {Brockman, Greg and Cheung, Vicki and Pettersson, Ludwig and Schneider, Jonas and Schulman, John and Tang, Jie and Zaremba, Wojciech},
	month = jun,
	year = {2016},
	note = {arXiv:1606.01540 [cs.LG]},
}

@inproceedings{chuangDebiasedContrastiveLearning2020,
	title = {Debiased {Contrastive} {Learning}},
	volume = {33},
	url = {https://papers.neurips.cc/paper_files/paper/2020/hash/63c3ddcc7b23daa1e42dc41f9a44a873-Abstract.html},
	urldate = {2026-08-08},
	booktitle = {Advances in {Neural} {Information} {Processing} {Systems}},
	publisher = {Curran Associates, Inc.},
	author = {Chuang, Ching-Yao and Robinson, Joshua and Lin, Yen-Chen and Torralba, Antonio and Jegelka, Stefanie},
	year = {2020},
	pages = {8765--8775},
}

@inproceedings{khoslaSupervisedContrastiveLearning2020,
	title = {Supervised {Contrastive} {Learning}},
	volume = {33},
	url = {https://proceedings.neurips.cc/paper/2020/hash/d89a66c7c80a29b1bdbab0f2a1a94af8-Abstract.html},
	urldate = {2026-08-07},
	booktitle = {Advances in {Neural} {Information} {Processing} {Systems}},
	publisher = {Curran Associates, Inc.},
	author = {Khosla, Prannay and Teterwak, Piotr and Wang, Chen and Sarna, Aaron and Tian, Yonglong and Isola, Phillip and Maschinot, Aaron and Liu, Ce and Krishnan, Dilip},
	year = {2020},
	pages = {18661--18673},
}

@article{bignoldHumanEngagementProviding2023,
	title = {Human {Engagement} {Providing} {Evaluative} and {Informative} {Advice} for {Interactive} {Reinforcement} {Learning}},
	volume = {35},
	issn = {0941-0643, 1433-3058},
	url = {http://arxiv.org/abs/2009.09575},
	doi = {10.1007/s00521-021-06850-6},
	number = {25},
	urldate = {2026-07-20},
	journal = {Neural Computing and Applications},
	author = {Bignold, Adam and Cruz, Francisco and Dazeley, Richard and Vamplew, Peter and Foale, Cameron},
	month = sep,
	year = {2023},
	note = {arXiv:2009.09575 [cs.AI]},
	pages = {18215--18230},
}

@misc{chenSimpleFrameworkContrastive2020,
	title = {A {Simple} {Framework} for {Contrastive} {Learning} of {Visual} {Representations}},
	url = {http://arxiv.org/abs/2002.05709},
	doi = {10.48550/arXiv.2002.05709},
	urldate = {2026-07-19},
	publisher = {arXiv},
	author = {Chen, Ting and Kornblith, Simon and Norouzi, Mohammad and Hinton, Geoffrey},
	month = jul,
	year = {2020},
	note = {arXiv:2002.05709 [cs.LG]},
}

@misc{stookeDecouplingRepresentationLearning2021,
	title = {Decoupling {Representation} {Learning} from {Reinforcement} {Learning}},
	url = {http://arxiv.org/abs/2009.08319},
	doi = {10.48550/arXiv.2009.08319},
	language = {en},
	urldate = {2026-07-19},
	publisher = {arXiv},
	author = {Stooke, Adam and Lee, Kimin and Abbeel, Pieter and Laskin, Michael},
	month = may,
	year = {2021},
	note = {arXiv:2009.08319 [cs.LG]},
}

@inproceedings{yaratsReinforcementLearningPrototypical2021,
	title = {Reinforcement {Learning} with {Prototypical} {Representations}},
	issn = {2640-3498},
	url = {https://proceedings.mlr.press/v139/yarats21a.html},
	language = {en},
	urldate = {2026-07-19},
	booktitle = {Proceedings of the 38th {International} {Conference} on {Machine} {Learning}},
	publisher = {PMLR},
	author = {Yarats, Denis and Fergus, Rob and Lazaric, Alessandro and Pinto, Lerrel},
	month = jul,
	year = {2021},
	pages = {11920--11931},
}

@book{murphyProbabilisticMachineLearning2022,
	address = {Cambridge, UNITED STATES},
	title = {Probabilistic {Machine} {Learning}: {An} {Introduction}},
	isbn = {978-0-262-36930-5},
	shorttitle = {Probabilistic {Machine} {Learning}},
	url = {http://ebookcentral.proquest.com/lib/unsw/detail.action?docID=6884472},
	urldate = {2026-07-19},
	publisher = {MIT Press},
	author = {Murphy, Kevin P.},
	year = {2022},
}

@article{bignoldEvaluationMethodologyInteractive2021,
	title = {An {Evaluation} {Methodology} for {Interactive} {Reinforcement} {Learning} with {Simulated} {Users}},
	volume = {6},
	issn = {2313-7673},
	url = {https://pmc.ncbi.nlm.nih.gov/articles/PMC7985787/},
	doi = {10.3390/biomimetics6010013},
	number = {1},
	urldate = {2026-07-19},
	journal = {Biomimetics},
	author = {Bignold, Adam and Cruz, Francisco and Dazeley, Richard and Vamplew, Peter and Foale, Cameron},
	month = feb,
	year = {2021},
	pages = {13},
}

@book{suttonReinforcementLearningSecond2018,
	address = {Cambridge, UNITED STATES},
	title = {Reinforcement {Learning}, {Second} {Edition}: {An} {Introduction}},
	isbn = {978-0-262-35270-3},
	shorttitle = {Reinforcement {Learning}, {Second} {Edition}},
	url = {http://ebookcentral.proquest.com/lib/unsw/detail.action?docID=6260249},
	urldate = {2026-03-09},
	publisher = {MIT Press},
	author = {Sutton, Richard S. and Barto, Andrew G.},
	year = {2018},
}

@book{plaatDeepReinforcementLearning2022,
	address = {Singapore},
	title = {Deep {Reinforcement} {Learning}},
	copyright = {https://www.springer.com/tdm},
	isbn = {978-981-19-0637-4 978-981-19-0638-1},
	url = {https://link.springer.com/10.1007/978-981-19-0638-1},
	doi = {10.1007/978-981-19-0638-1},
	language = {en},
	urldate = {2026-07-19},
	publisher = {Springer Nature},
	author = {Plaat, Aske},
	year = {2022},
}

@article{nguyenBroadPersistentAdvisingApproach2023,
	title = {Towards a {Broad}-{Persistent} {Advising} {Approach} for {Deep} {Interactive} {Reinforcement} {Learning} in {Robotic} {Environments}},
	volume = {23},
	copyright = {http://creativecommons.org/licenses/by/3.0/},
	issn = {1424-8220},
	url = {https://www.mdpi.com/1424-8220/23/5/2681},
	doi = {10.3390/s23052681},
	language = {en},
	number = {5},
	urldate = {2026-03-10},
	journal = {Sensors},
	publisher = {Multidisciplinary Digital Publishing Institute},
	author = {Nguyen, Hung Son and Cruz, Francisco and Dazeley, Richard},
	month = mar,
	year = {2023},
}

@misc{bengioRepresentationLearningReview2012,
	title = {Representation {Learning}: {A} {Review} and {New} {Perspectives}},
	shorttitle = {Representation {Learning}},
	url = {https://arxiv.org/abs/1206.5538v3},
	language = {en},
	urldate = {2026-04-03},
	journal = {arXiv.org},
	author = {Bengio, Yoshua and Courville, Aaron and Vincent, Pascal},
	month = jun,
	year = {2012},
}

@article{moreiraDeepReinforcementLearning2020,
	title = {Deep {Reinforcement} {Learning} with {Interactive} {Feedback} in a {Human}-{Robot} {Environment}},
	volume = {10},
	issn = {2076-3417},
	url = {http://arxiv.org/abs/2007.03363},
	doi = {10.3390/app10165574},
	number = {16},
	urldate = {2026-04-03},
	journal = {Applied Sciences},
	author = {Moreira, Ithan and Rivas, Javier and Cruz, Francisco and Dazeley, Richard and Ayala, Angel and Fernandes, Bruno},
	month = aug,
	year = {2020},
	note = {arXiv:2007.03363 [cs]},
	pages = {5574},
}

@misc{cruzSurveyInteractiveReinforcement2021,
	title = {A {Survey} on {Interactive} {Reinforcement} {Learning}: {Design} {Principles} and {Open} {Challenges}},
	shorttitle = {A {Survey} on {Interactive} {Reinforcement} {Learning}},
	url = {http://arxiv.org/abs/2105.12949},
	doi = {10.48550/arXiv.2105.12949},
	urldate = {2026-04-03},
	publisher = {arXiv},
	author = {Cruz, Christian Arzate and Igarashi, Takeo},
	month = may,
	year = {2021},
	note = {arXiv:2105.12949 [cs]},
}

@misc{srinivasCURLContrastiveUnsupervised2020,
	title = {{CURL}: {Contrastive} {Unsupervised} {Representations} for {Reinforcement} {Learning}},
	shorttitle = {{CURL}},
	url = {http://arxiv.org/abs/2004.04136},
	doi = {10.48550/arXiv.2004.04136},
	urldate = {2026-03-12},
	publisher = {arXiv},
	author = {Srinivas, Aravind and Laskin, Michael and Abbeel, Pieter},
	month = sep,
	year = {2020},
	note = {arXiv:2004.04136 [cs]},
}

@misc{echchahedSurveyStateRepresentation2025,
	title = {A {Survey} of {State} {Representation} {Learning} for {Deep} {Reinforcement} {Learning}},
	url = {http://arxiv.org/abs/2506.17518},
	doi = {10.48550/arXiv.2506.17518},
	urldate = {2026-03-09},
	publisher = {arXiv},
	author = {Echchahed, Ayoub and Castro, Pablo Samuel},
	month = jun,
	year = {2025},
	note = {arXiv:2506.17518 [cs]},
}

@article{najarReinforcementLearningHuman2021,
	title = {Reinforcement {Learning} {With} {Human} {Advice}: {A} {Survey}},
	volume = {8},
	issn = {2296-9144},
	shorttitle = {Reinforcement {Learning} {With} {Human} {Advice}},
	url = {https://www.frontiersin.org/journals/robotics-and-ai/articles/10.3389/frobt.2021.584075/full},
	doi = {10.3389/frobt.2021.584075},
	language = {English},
	urldate = {2026-03-09},
	journal = {Frontiers in Robotics and AI},
	publisher = {Frontiers},
	author = {Najar, Anis and Chetouani, Mohamed},
	month = jun,
	year = {2021},
}
\bibliographystyle{named}

\end{document}